# Extraction of Skin Lesions from Non-Dermoscopic Images Using Deep Learning

Mohammad H. Jafari, Ebrahim Nasr-Esfahani, Nader Karimi, S.M. Reza Soroushmehr, Shadrokh Samavi,
Kayvan Najarian

*Abstract*— **Melanoma is amongst most aggressive types of cancer. However, it is highly curable if detected in its early stages. Prescreening of suspicious moles and lesions for malignancy is of great importance. Detection can be done by images captured by standard cameras, which are more preferable due to low cost and availability. One important step in computerized evaluation of skin lesions is accurate detection of lesion's region, i.e. segmentation of an image into two regions as lesion and normal skin. Accurate segmentation can be challenging due to burdens such as illumination variation and low contrast between lesion and healthy skin. In this paper, a method based on deep neural networks is proposed for accurate extraction of a lesion region. The input image is preprocessed and then its patches are fed to a convolutional neural network (CNN). Local texture and global structure of the patches are processed in order to assign pixels to lesion or normal classes. A method for effective selection of training patches is used for more accurate detection of a lesion's border. The output segmentation mask is refined by some post processing operations. The experimental results of qualitative and quantitative evaluations demonstrate that our method can outperform other state-of-the-art algorithms exist in the literature.**

*Index Terms*— **Convolutional neural network, deep learning, medical image segmentation, melanoma, skin cancer.**

## I. INTRODUCTION

SKIN cancer is the most prevailing type of cancer in the USA. While only about 1% of the diagnosed cases of skin cancer are melanoma, they cause 75% of related deaths [1]. It is estimated that 76,000 new cases of melanoma will be detected in the USA in 2016. It kills about 10,000 people in each year, i.e. nearly one person dies of melanoma in every hour [2]. Despite of its great danger, melanoma is highly treatable in its early stages [3]. The five-year survival rate for melanoma is 98% for those in local stage, 63% in regional stage and 17% in distant stage [1]. This means treatment chance drops significantly as the detection is postponed. Noticing the importance of early detection, dermatologists have developed various metrics for evaluation of skin lesions for malignancy [3]. ABCD criterion (asymmetry of the lesion, border irregularity, color variation, and diameter) [4], and the seven-point checklist [5] are two widely used methods that help differentiate between melanoma and non-melanoma (benign) lesions.

There is an active research trend for computerized prescreening of suspicious skin lesions for malignancy. A review of literature for this purpose is available in [6] and [7]. Computerized analysis of pigmented skin lesions usually comprises of the following two main steps performed on the input image of skin [7]: (I) preprocessing and segmentation, in which image enhancement and artifact reduction are done and the image is segmented into two regions as lesion and normal skin, (II) feature extraction and classification, where some dermatologically important aspects of the extracted lesion are assessed to evaluate its malignancy risk [8]. Computerized analysis of skin lesions is usually performed on two categories of input images [7] as (I) dermoscopic (i.e. microscopic images) and (II) non-dermoscopic (i.e. macroscopic or clinical images). Dermoscopic images are produced by dermatoscope, a special instrument that facilitates the procedure of inspection by magnifying the lesion and providing uniform illumination. Dermoscopic images are not easily available and less than 50% of the dermatologists in the US utilize dermatoscopy [9]. On the other hand, non dermoscopic images have the advantage of being easily accessible. These images are captured by conventional user grade cameras and produce what is seen by naked eye. This can be a tool for non-specialists to evaluate the risk of suspicious skin lesions [8, 10-11] and can be used as a means of priority evaluation and patient scheduling for appointments. Contrary to dermoscopic images, clinical images may have illumination variations, contain less detailed information, and have a lower contrast. These factors can affect the process of segmentation and feature extraction.

Mohammad H. Jafari, Ebrahim Nasr-Esfahani, and Nader Karimi are with the Department of Electrical and Computer Engineering, Isfahan University of Technology, Isfahan 84156-83111, Iran.

S.M. Reza Soroushmehr is with the Michigan Center for Integrative Research in Critical Care, and also the Emergency Medicine Department, University of Michigan, Ann Arbor, U.S.A.

Shadrokh Samavi is with the Department of Electrical and Computer Engineering, Isfahan University of Technology, Isfahan 84156-83111, Iran. He is also with the Department of Emergency Medicine, University of Michigan, Ann Arbor, U.S.A.

Kayvan Najarian is with the Department of Computational Medicine and Bioinformatics, Department of Emergency Medicine, and the Michigan Center for Integrative Research in Critical Care, University of Michigan, Ann Arbor, U.S.A.

The first step, in an automatic evaluation of pigmented skin lesions, is accurate detection of lesion region. This stage directly affects the usability of extracted features in the next step. Dermatologically important aspects such as lesion asymmetry and border irregularity are mainly derived from the segmentation mask. Various segmentation methods exist for this purpose, which can be classified into three groups [12] as active contours, region merging [13] and thresholding. A review of existing methods for segmentation of dermoscopic images is given in [12] and [14]. Here we are only considering methods that use non-dermoscopic images. Because of the mentioned difficulties, such as illumination variation, presence of artifacts, and low contrast between lesion and normal skin, some methods are specifically proposed for lesion segmentation in these images [15-18]. Some of the methods consider color information [15-17]. Work of [15] considers the difference between the lesion and normal skin parts in red channel of RGB color space and performs thresholding to segment the image. Method of [16] performs on multiple color channels and [17] utilizes a discriminating 3-channel space that is calculated by principle component analysis (PCA). However, color based methods fail to notice the textural information such as smoothness or raggedness that could aid the segmentation procedure. In the method proposed in [18], a set of sparse texture distributions representing the input image is learned. Then the distinctiveness of texture of the lesion as compared to the normal skin is considered for the segmentation. Incorporation of textual information usually improves segmentation performance, meanwhile these methods might be misguided by complexity of texture, color pattern and structure of the lesion and skin.

Skin images, taken by standard cameras, usually have uneven lightening caused by reflection of light from the skin surface. Most of the segmentation algorithms are vulnerable to this misleading factor and they make incorrect segmentation. The mentioned methods [15-18] comprise a preprocessing step that deals with illumination variations. A review of various methods, for illumination modelling and correction, is given in [19]. Moreover, usually some preprocessing methods are performed in order to handle bothersome artifacts such as skin hair and specular reflections [7].

Recently deep learning approaches have shown an emerging growth and have resulted in promising improvements in different fields of pattern recognition. A review of literature is given in [20] and [21]. Convolutional neural networks (CNNs) are widely used approaches in deep learning and are applicable in diverse image processing problems. Works of [22-25] are recent researches using deep neural networks for segmentation of medical images. In [22] and [23] CNN is used for segmentation of vessels in fundus images and X-ray angiogram. Works of [24] and [25] have used CNNs for segmentation of brain tumors. Moreover a CNN in [26] is applied for classification of suspicious skin lesions. CNN can be used as an effective tool for automatic extraction of discriminative features. Furthermore, the architecture of CNN can be of great importance when adapting to a specific purpose. In some applications, combination of local texture and global context (local view and a broader look at the same spot) could significantly improve performance of segmentation, as it is used in [25] and [27]. It is shown that CNNs are among powerful approaches for effective feature extraction and classification. This could be a motivation to consider such a network for segmentation of lesions from skin images. The extraction of color and texture features that are discriminating between lesion and normal skin could be performed by the CNN. The CNN would learn the features based on training samples. The preliminary idea of such a method is presented in [28].

In this paper we perform segmentation of skin lesions using CNNs. Firstly the input image is preprocessed with the aim of ridding the image from artifacts that can affect segmentation. Afterward, around each pixel a local and a global patch are considered. A local patch would process information of local texture and the corresponding global patch would reveal the general structure around that pixel. Combination of local and global information would enhance the accuracy of border detection and improves the robustness of our method against illumination variations. About 420,000 patches are acquired and used for the training of the CNN. Also, to enhance the accuracy of the border detection, a method is proposed for effective selection of training patches. The experimental results show that our proposed segmentation method is superior in terms of accuracy in comparison with other state-of-the-art algorithms.

The remainder of this paper is organized as follows. Section II explains the proposed method for segmentation of skin lesions in details. In Section III, the effects of combination of local and global patches are discussed. Experimental results are given in Section IV and finally Section V concludes the paper.

## II. SKIN LESION SEGMENTATION

In this section, our proposed method for extraction of skin lesion from non-dermoscopic images is explained. Figure 1 presents an overview of our method, where different steps for the training of the CNN and the testing procedure are shown. In the following all of these steps are discussed in details.

### A. Pre Processing

The images of skin captured by standard cameras usually contain artifacts such as light scattering, caused by specular reflection of the light from skin surface, and skin parts such as hair, that could adversely affect the segmentation performance. In our method to deal with these factors, the input image is firstly preprocessed by applying a guided filter [29] that is amongst powerful edge preserving smoothing operators. It can also reduce irrelevant artifacts while maintaining important information for the sake of segmentation. Guided filtering of the input image would clarify the skin image from disturbing elements, while causing minimum distortion on lesion's border.

Both training and test images are preprocessed by a guided filter. Sample results of this step are shown in Fig. 2. The guided filter takes one input image for filtering and uses another one to guide the filtering process. In our work, the input image itself is used as the guidance image. As can be seen in Fig. 2, by applying this step the images are enhanced for segmentation and effects of bothersome factors, which could mislead the CNN's learning process, are reduced. It is done while preserving the shape of the lesion and the boundary between the lesion and normal skin. The boundary is of great importance for accurate segmentation.

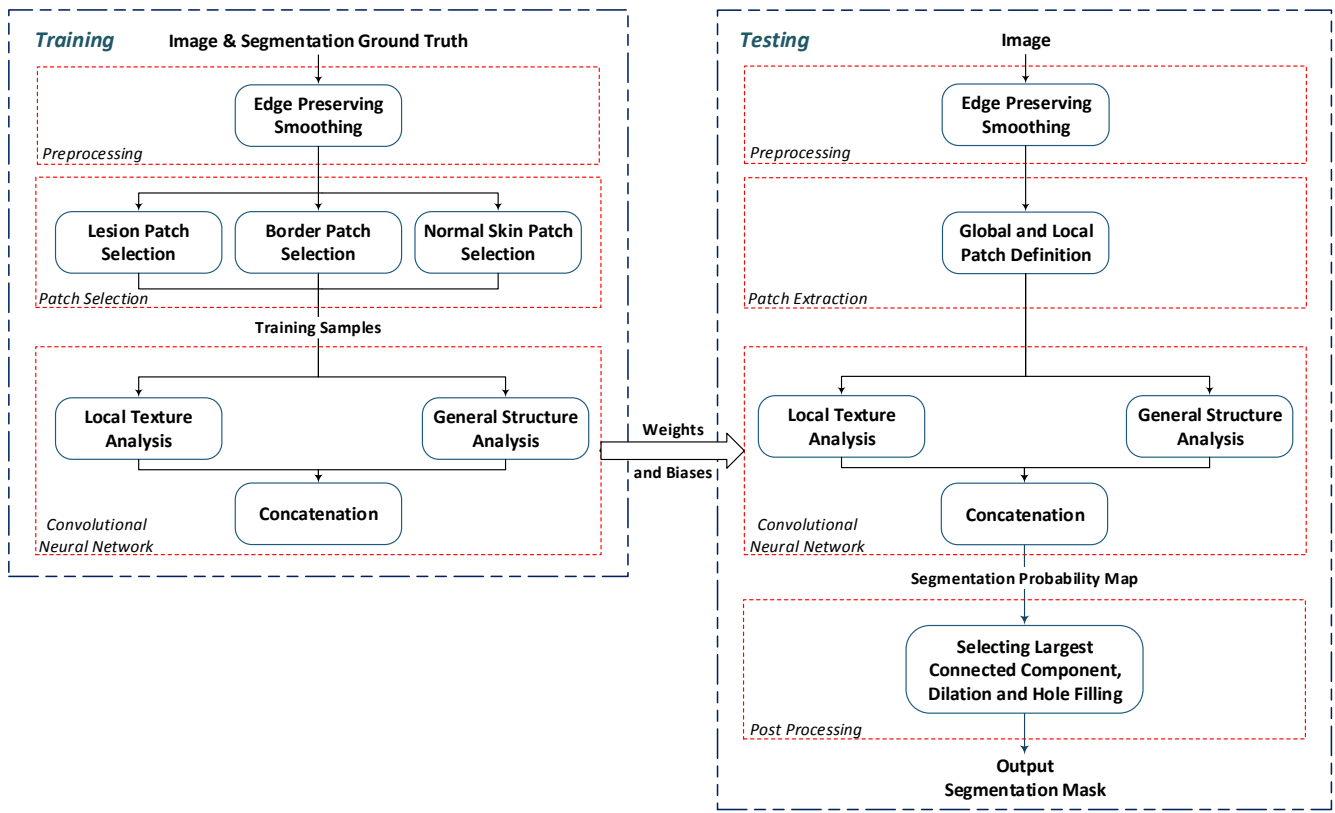

Fig. 1. Block diagram of the proposed method.

### B. Patch Extraction

In order to segment the preprocessed image, a pixel classification is performed. For this purpose, a window as a patch is considered for each pixel, where the pixel would be in the center of the patch. For each pixel a local and a global patch are considered. A small size window or patch reveals the local texture around the pixel that is helpful for the segmentation purpose. Usually there exist some clear differences between local texture of the lesion and that of the normal skin. Moreover, a local patch would be needed for precise detection of lesion's border. Hence a local $31 \times 31$ window is considered around each pixel of the image. On the other hand, a large size patch would show the general structure of the region around the pixel. A global patch is not misled by local irrelevant information, such as artifacts and illumination variations. In our work, a window of size $201 \times 201$ is considered around each pixel. This global patch is then resized to $31 \times 31$. Therefore, we would have a zoomed out view of a location where the intended pixel is at its center. Input images are resized to $400 \times 600$ pixels; hence each image would be made up of 240,000 pairs of local patch and its corresponding global patch. In case that a part of a patch falls outside of the image, the image is padded by replication of the image border pixels. The padded region is smoothed by applying a mean filter, since usually in input images a lesion is placed in center and the mean value of border pixels would be an estimation of the normal skin color. The extracted patches are then fed to a CNN to segment the skin lesion. The architecture of the CNN is explained in the next section.

### C. CNN Architecture

Each pixel will eventually be labeled as normal or lesion. The pairs of patches are applied to a CNN with the architecture that is shown in Fig. 3. Local and global patches are processed simultaneously in two parallel paths of the CNN. The two CNN paths have similar layers and configurations. The order of layers in these two paths is as *Conv1*, *MaxPool1*, *Conv2*, and *MaxPool2*. *Conv1* and *Conv2* denote convolutional layers, where kernel size in *Conv1* is $6 \times 6 \times 3$ and is $5 \times 5 \times 60$ in *Conv2*. There are 60 feature maps in each convolve layer, which means each convolve layer can detect 60 different kinds of features across its input patch. These features are learned by the CNN as to be discriminative between lesion and normal skin regions. The hyper parameters in the network such as size of input patches and kernels and number of feature maps are set by test and trial.

In CNNs, the convolve layers are usually followed by pooling layers. This means that output of a convolve layer would be abstracted by selection of candidate values. The abstraction would reduce the number of variables in the network by simplifying the information. As can be seen in Fig. 3, in the used CNN two layers of max pooling as *MaxPool1* and *Maxpool2* are included after first and second convolve layer. In a max-pooling layer, output of previous layer is summarized through selection of max value in a window sliding on the image. Kernel size in MaxPool1 and MaxPool2 is $2 \times 2$ and $3 \times 3$ and the stride length of the sliding windows for these layers is set to 2 and 3 respectively.

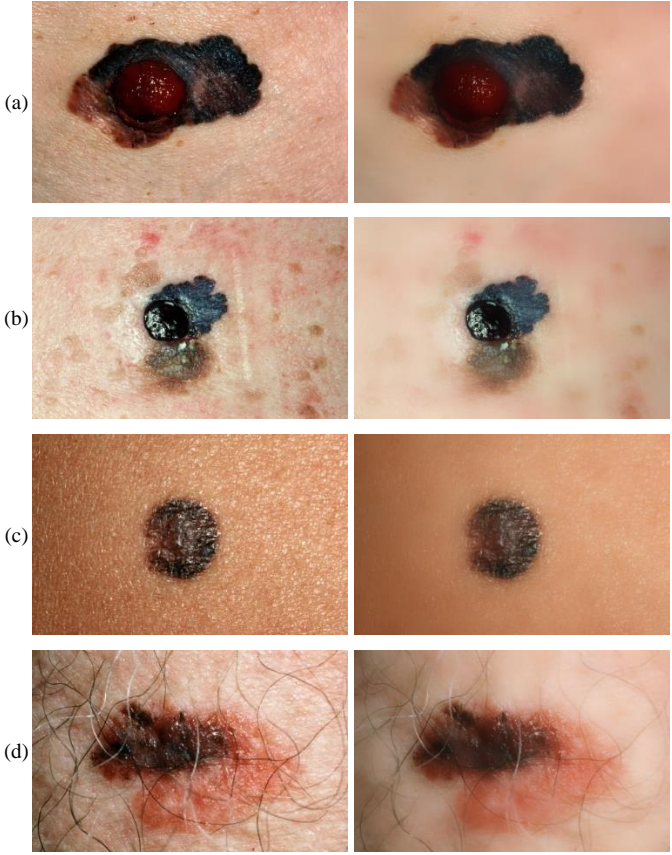

Fig. 2. Left column are input images. Right column are preprocessed images by applying the guided filter.

Outputs of the two parallel networks, which have processed local and global patches, are fed into a fully connected concatenation layer with 500 neurons. This layer combines the extracted information of local texture and global structure around the pixel. Finally, this layer is followed by an output *softmax* layer with 2 neurons. Output of the network indicates a probability of membership of each pixel to lesion or normal skin groups.

### D. Post Processing

By feeding an image to the CNN, using patches corresponding to each pixel, a probability map is obtained. In this map, each point has a value $0 \leq P(x,y) \leq 1$, which is the probability of the corresponding pixel's membership in the lesion region. In order to produce a binary mask for the lesion region, a threshold as $\tau$ is used where all points with a value of probability more than $\tau$, i.e. $P(x,y) > \tau$, will be labeled as a lesion class. The map is labeled 1 if the pixel is lesion and 0 if the pixel is in a normal skin region. As the final step, the output mask is further refined by selecting the largest connected component, applying a morphological dilation operation and a hole-filling procedure. In our method, the threshold $\tau$ is chosen to be 0.6. The segmentation results of our method and quantitative comparisons are presented in Section IV. In the following, the used method for effective selection of training patches is discussed.

### E. Selection of Training Patches

One important factor that affects the learning procedure of the CNN is the proper selection of the training patches. In our method, a subset of images in the dataset is used for the training. The question that arises here is that should the CNN be trained using all the available samples from training images?

On one side, applying all of the patches to the CNN would be cumbersome. It would create a huge training dataset that consists of about 20 million patches ($240000 *$ number of training images). This would cause storage and computation complexities. Also it should be noticed that a high portion of these patches have similar information. On the other side, random selection of a small portion of all available patches might cause loss of some important information. The ignored samples might have some critical data for CNN training and hence it could reduce the learning accuracy. For example, by pure random selection many of the patches may be selected from the non-lesion regions. This could cause loss of information from the lesion part, or vice versa. As a result, it is necessary to choose a portion of patches as candidates that represent the essential information of the image for the lesion segmentation.

A solution to this problem is a smart selection of a small set of training patches. For this aim, we use the ground truth segmentation for the selection of sample patches. In our method, 1/3 of the patches are randomly selected from the lesion region. These patches are shown by red dots in Fig. 4 (d). Also, 1/3 of patches are extracted from normal skin parts (shown by blue markers in Fig. 4 (d)). We also noticed that the area close to the border of the lesion could be a challenging region for CNN to detect. By guiding the CNN to learn border region, the segmentation performance could be improved. Hence rest of the patches, i.e. last 1/3, are selected randomly from the border area (shown by cyan markers in Fig. 4 (d)). For this aim, the border region is defined by a morphological operation as:

$$Margin = (GT \oplus SE) - (GT \ominus SE), \quad (1)$$

where $\oplus$ means morphological dilation and $\ominus$ is morphological erosion. GT denotes segmentation ground truth and SE is a disk structure element with $radius = 15$.

The border region is shown for a sample training image in Fig. 4 (c). Also, the lesion and normal skin regions can be seen in Fig. 4 (b). This selection approach for the training data helps reduction of the required number of learning patches, while effective information for lesion segmentation can be obtained.

Selection of training patches from the border area, as a challenging region for segmentation, could noticeably enhance the performance of trained CNN. Border contours are shown in Fig. 5 (b) where a CNN is trained by random selection of patches from lesion and non-lesion regions. These outputs are compared with those in Fig. 5 (c) where a CNN with same configurations is trained with the mentioned three groups of patches (i.e. smart selection from the border, lesion, and normal skin regions). As is shown in Fig. 5, training the CNN by border patches could reduce the number of falsely classified pixels and better extracts the border curvature details.

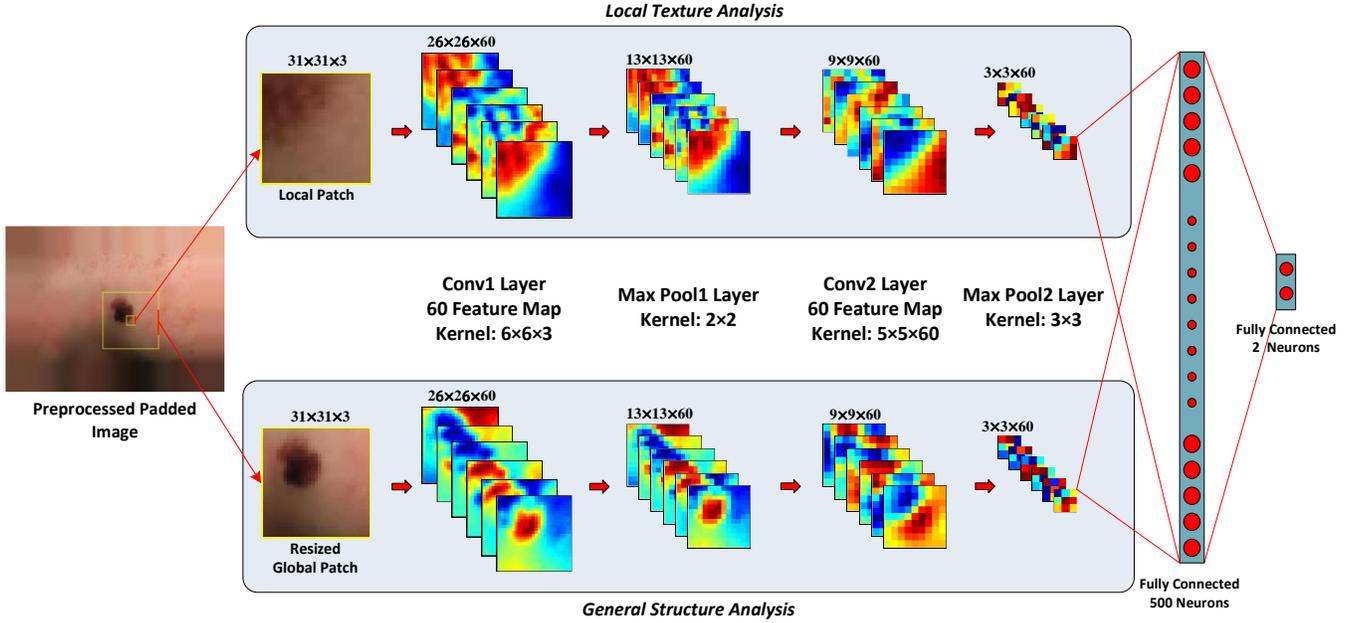

Fig. 3. Architecture of the proposed network.

## III. ANALYZING EFFECTS OF LOCAL AND GLOBAL PATCHES

The final output in our method is calculated by concatenating the results of processing local and global patches. In order to measure the effectiveness of combination of two sources of information, the output probability maps of independently applying each of the two parallel networks (local and global) are obtained. For this aim, two experiments are done, where two networks are configured. One of the experiments is only trained and performed on local patches (local texture analysis), and the other CNN is only based on global patches (general structure analysis).

Some sample probability maps produced by local and global networks are shown in Fig. 6. As can be seen, a network that is only based on local information could be distracted by local factors such as illumination variations. On the contrary, skin lesions can be properly detected in a general view, but details of border's shape, which are of great importance for medical diagnosis, are not extracted accurately. This is due to the fact that the detailed information is not available in global patches.

With the combination of information in local and global patches the performance of segmentation is increased considerably. Global information could prevent the CNN from falling in the pitfall of misguiding information such as uneven illumination or irrelevant local texture. Moreover, local information could assist the procedure of exact border detection. This can be seen in Fig. 6, where sample probability maps produced by local and global networks are shown along with the final probability map and segmentation of our proposed method. The output is derived from the CNN that combines local and global information. As can be seen, by adding the information of local patches, the accuracy of segmentation contour enhances significantly and covers the lesion's border with higher precision. Moreover for quantitative comparison, numerical segmentation results of the mentioned methods are available in the following section.

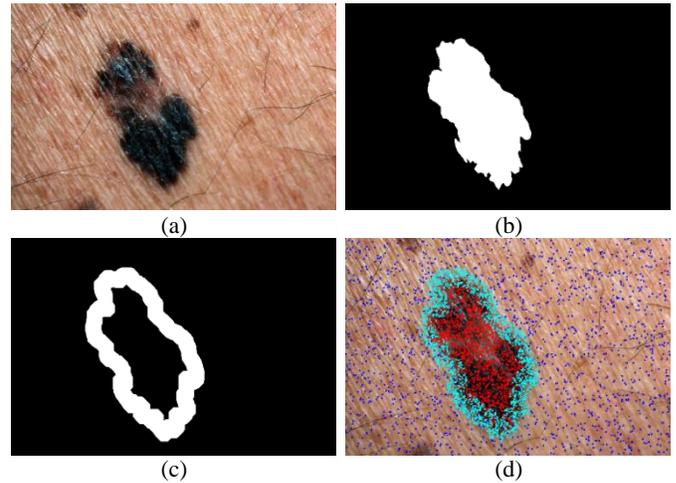

Fig. 4. Regions for random selection of training patches. (a) sample image for training, (b) lesion region (white) and normal skin (black), (c) margin of the lesion's border (white) and (d) center of 4,500 selected training patches, shown by red markers for lesion region, blue for normal skin and cyan for border margin.

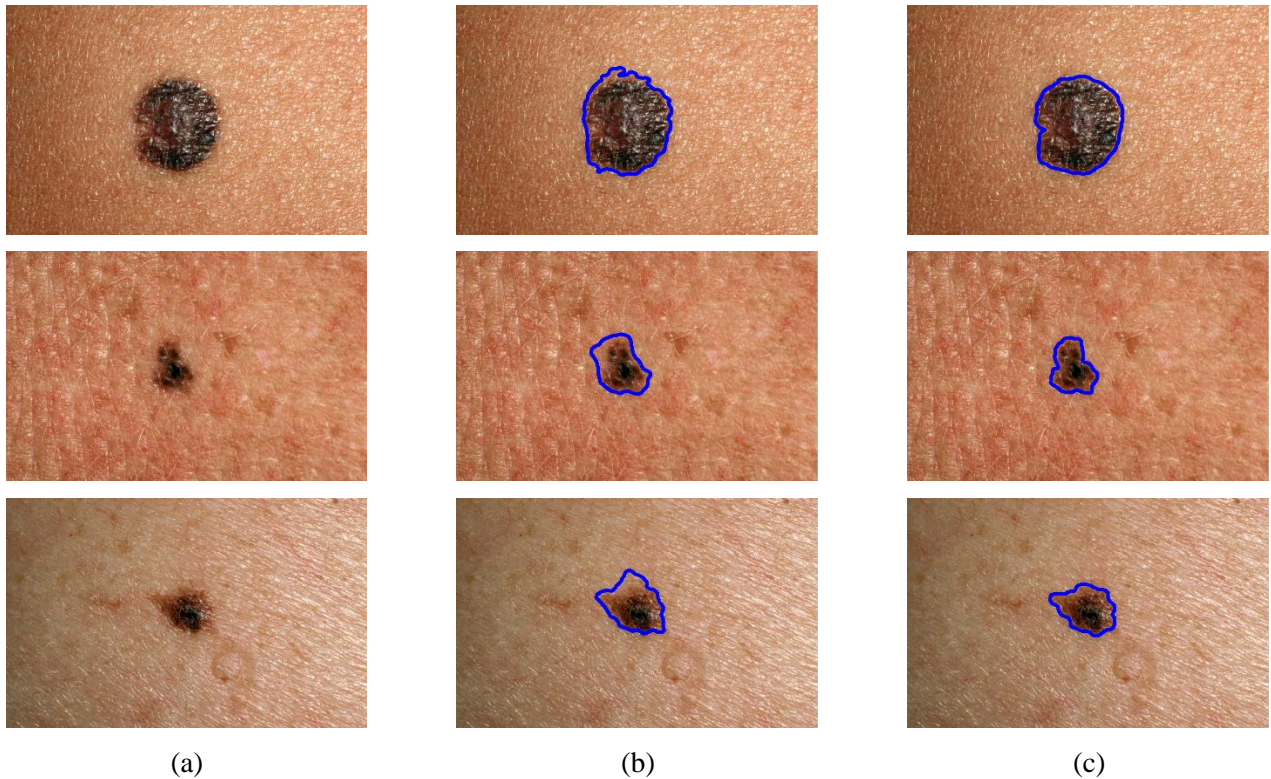

| (a) | (b) | (c) |

Fig. 5. Effects of training the CNN on border region, (a) sample test images, (b) results of CNN that is trained on patches from only lesion and normal skin and (c) results for adding selected border patches to training samples.

## IV. EXPERIMENTAL RESULTS

Experiments, done for qualitative and quantitative evaluation of our method, are presented in this section. For this aim, our method is tested on a dataset of skin lesion images from Dermquest database that is publically available with segmentation ground-truth [30]. This dataset consists of 126 digital images, where 66 of images are melanoma cases and the others are non-melanoma. The proposed method is implemented in Matlab and Caffe [31]. The experiment is done as a leave-p-out cross-validation. The dataset is randomly split into four groups with equal sizes. One group is left for test and the other three groups are used for training, i.e. the ratio of train to test is 75% to 25%. This Experiment is repeated four times for all of the test groups. The number of pair of patches selected from each training image is 4,500, where 1,500 of them are randomly chosen from lesion region, 1,500 patches are randomly selected from non-lesion region and finally the remaining 1,500 patches are in the vicinity of lesion's borders. Thus, a total of about 425,250 randomly selected patches are extracted and used as training samples of the CNN in each run. The solver method is stochastic gradient descent (SGD) and for weight initialization of the network Xavier algorithm is used. Bias values in the network are initially set to zero. Moreover, the neighborhood size of the guided filter in preprocessing stage is set to 100 and the structure element used for the dilation operation in post processing is a disk shape with $radius = 10$.

### A. CNN Learned Filters

To have an understanding of the learned discriminating features, we analyzed the learned kernels of the trained CNN. The used CNN comprises of two parallel networks that process local and global patches. The learned filters of the first convolutional layer of the local and global networks are shown in Fig. 7 (a) and (b). Furthermore, some sample outputs of activation layers, showing the results of convolving the input patches with the learned filters, are given in this figure. This would give an insight about the features that the CNN has noticed in skin images. As can be seen in Fig. 7, the learned CNN has a focus on regions that are relevant to the segmentation procedure. In some filtered patches, the lesion's boundary is highlighted. In addition, it can be seen in some convolved patches that the lesion and normal skin textures are clearly discriminated.

### B. Qualitative Evaluation

Visual results of the proposed method, for some challenging images, are presented in Fig. 8. The detected lesion area is shown by a blue contour. Also, results of our method are compared with TDLS [18] that is a state-of-the-art method for lesion extraction from digital images. Images of TDLS are obtained from [32]. Comparison is done according to the ground truth segmentations that exist for the images of the dataset.

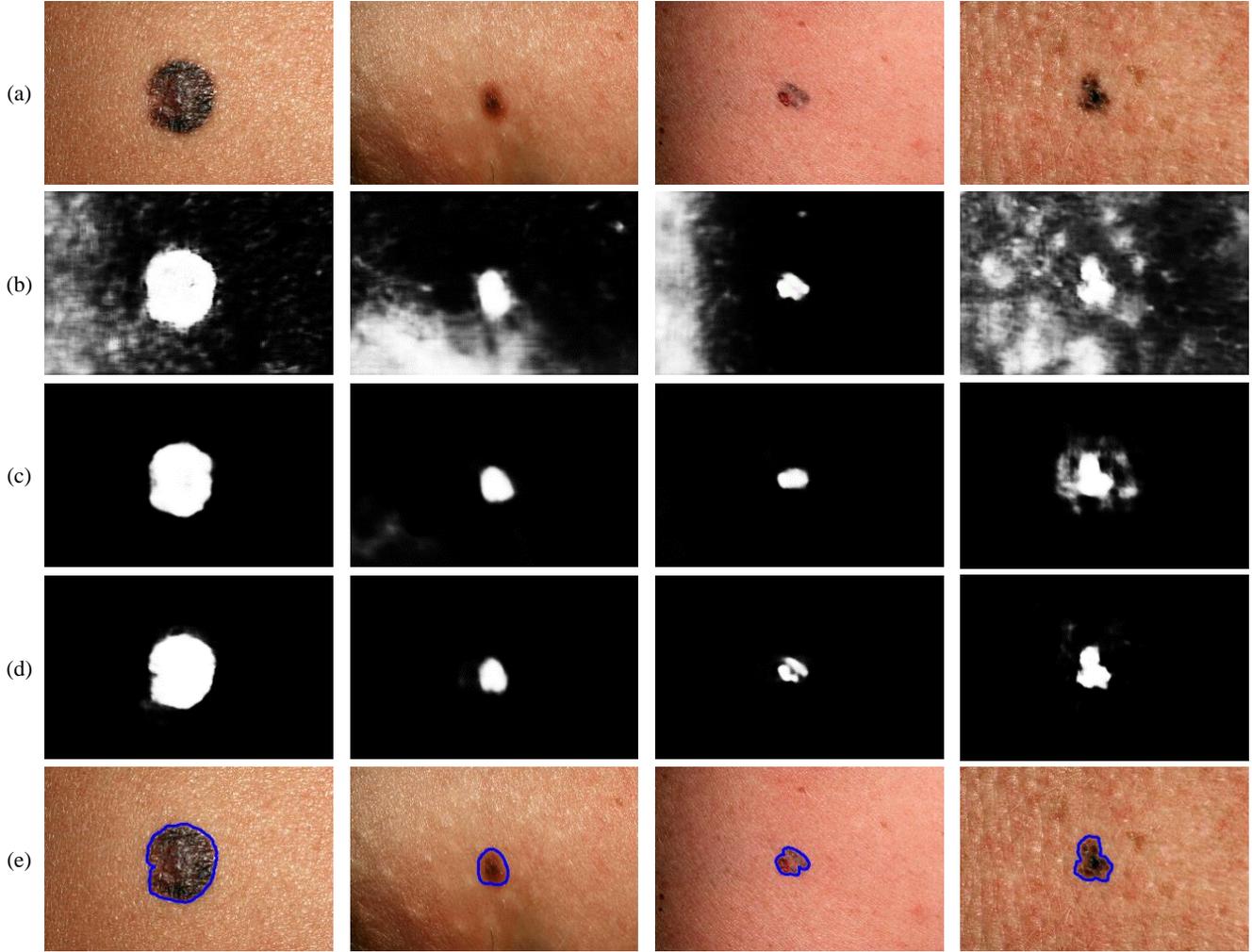

Fig. 6. Sample skin lesion segmentation results. (a) Input image, (b) probability map by local network (c) probability map by global network, (d) probability map by combination of local and global networks and (e) output segmentation contour after post processing.

As can be seen in Fig. 8, our method has a better performance in detection of the lesion region. Complex skin and lesion patterns and skin artifacts can misdirect any segmentation algorithm. In Fig. 8 (a) and (b), the TDLS method has a lower performance in extraction of lesions' borders and some pixels around the lesion's boundary are misclassified. In some samples, such as Fig. 8 (g), complicated texture of the lesion has misled the other method. Meanwhile, our method has extracted the lesion's border precisely even in challenging circumstances.

*C. Quantitative Evaluation*

Furthermore, performance of our method is measured quantitatively based on segmentation metrics. The segmentation of the lesion is considered as a classification problem. The pixels in the images should be classified into two groups of lesion and normal skin. Three metrics are used to evaluate the classification performance as:

$$sensitivity = \frac{TP}{TP + FN},$$

$$specificity = \frac{TN}{TN + FP},$$

$$accuracy = \frac{TP + TN}{TP + FN + TN + FP},$$

where TP is the number of lesion pixels classified correctly and FN counts the number of lesion pixels that are falsely classified. TN and FN denote the number of correctly and wrongly classified pixels of the normal skin respectively. Our proposed method takes advantage of combination of local and global patches. In Table I, the quantitative results of local and global nets are presented. As can be seen, combining these two sources of information would result in better segmentation performance.

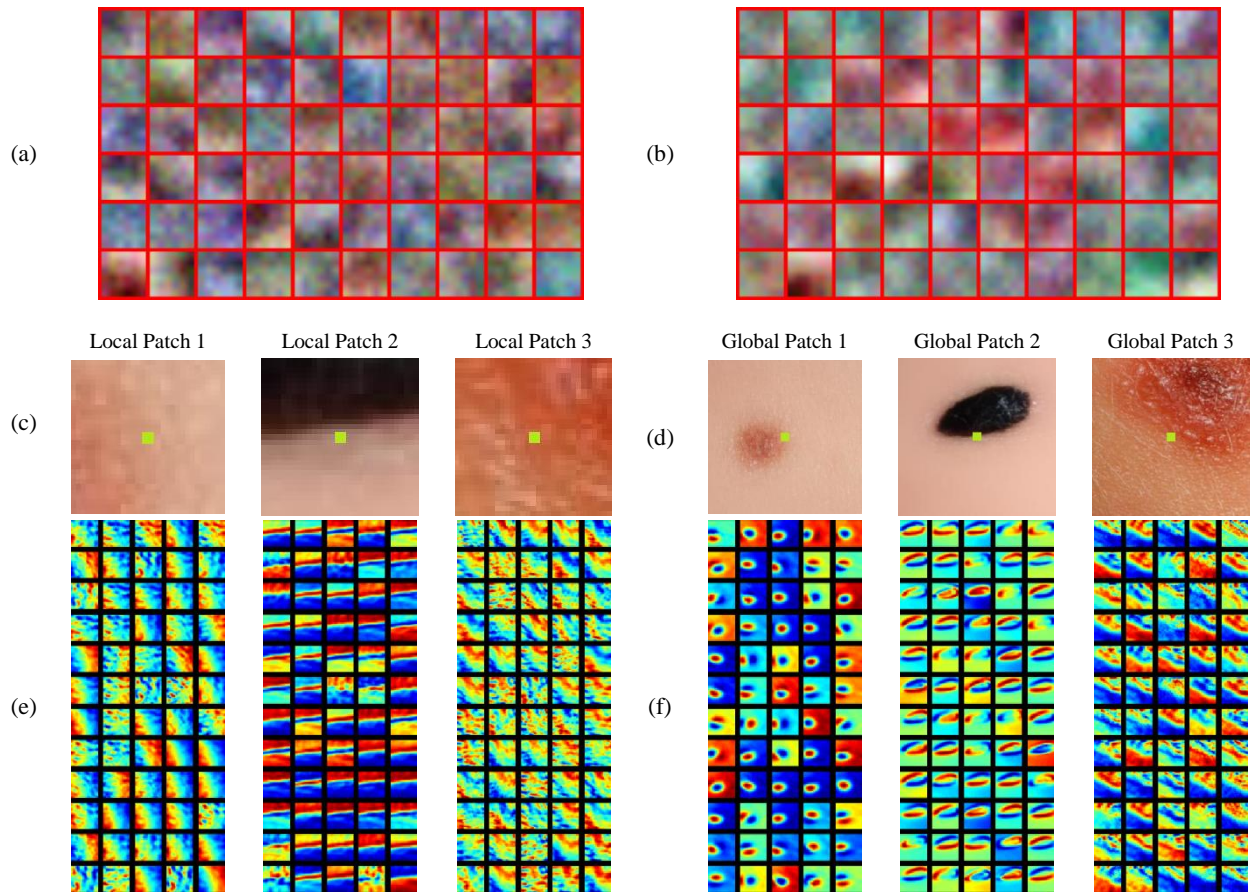

Fig. 7. Learned kernels and activation layers. Learned kernels of *Conv1* in (a) local and (b) global paths of the CNN. Three sample input patches where (c) are local and (d) are their corresponding global patches. Green dot is the center pixel. (e) Local and (f) global feature maps for the shown input patches.

Finally, our method is compared with five other methods that exist in the literature for skin lesion segmentation. The compared methods are namely as L-SRM [13], Otsu-R [15], Otsu-RGB [16], Otsu-PCA [17] and TDLS [18]. The numerical evaluation of these methods based on mentioned metrics on the same dataset is reported from [18].

The results are compared in Tables II to IV, where tests are performed on melanoma, non-melanoma and finally over all images of the dataset. As can be seen, our method has achieved a better score in sensitivity and accuracy of segmentation in comparison with other state-of-the-art methods. Also our method has the second highest specificity tested on melanoma cases. Otsu-PCA [17] has a better specificity in the melanoma group with a narrow lead of 0.7 percent; however our result in sensitivity leads over that method's score by 13.3 percent.

TABLE I
QUANTITATIVE MEASUREMENT OF SEGMENTATION PERFORMANCE BASED ON LOCAL PATCHES, GLOBAL PATCEHS AND THEIR COMBINATION, FOR ALL OF THE IMAGES IN THE DATASET. BEST RESULTS ARE BOLDED.

| Segmentation Algorithm | Segmentation Performance | | |
|---|---|---|---|
| | Sensitivity | Specificity | Accuracy |
| Local Texture | 89.5% | 95.6% | 95.2% |
| General Structure | 94.6% | 98.9% | 98.5% |
| Proposed Method (Combination) | **95.2%** | **99.0%** | **98.7%** |

TABLE II
QUANTITATIVE COMPARISON OF LESION SEGMENTATION RESULTS FOR MELANOMA CASES. BEST RESULTS ARE BOLDED.

| Segmentation Algorithm | Segmentation Performance | | |
|---|---|---|---|
| | Sensitivity | Specificity | Accuracy |
| L-SRM [13] | 90.0% | 92.5% | 92.1% |
| Otsu-R [15] | 87.4% | 91.5% | 90.3% |
| Otsu-RGB [16] | 92.2% | 85.5% | 85.0% |
| Otsu-PCA [17] | 81.2% | **99.5%** | 97.6% |
| TDLS [18] | 90.8% | 98.8% | 97.9% |
| Proposed Method | **94.5%** | 98.8% | **98.5%** |

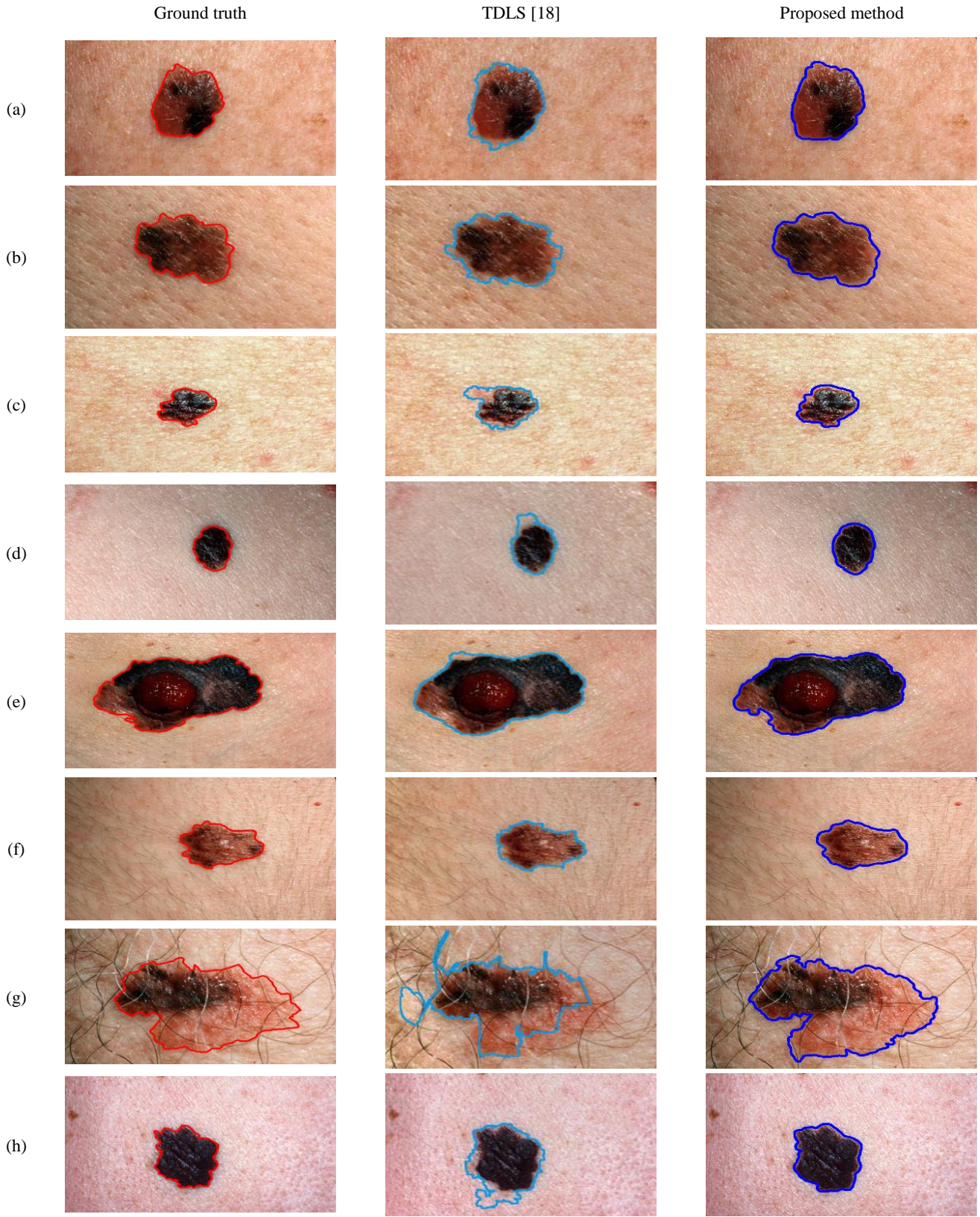

Fig. 8. Sample skin lesion segmentation results. Left column: input images superimposed by segmentation ground-truth (red line). Middle column: results of TDLS method [18]. Right column: masks resulted from our method.

TABLE III
QUANTITATIVE COMPARISON OF LESION SEGMENTATION RESULTS FOR NONMELANOMA CASES. BEST RESULTS ARE BOLDED.

| Segmentation Algorithm | Segmentation Performance | | |
| --- | --- | --- | --- |
| | Sensitivity | Specificity | Accuracy |
| L-SRM [13] | 88.7% | 93.0% | 92.6% |
| Otsu-R [15] | 87.3% | 78.7% | 78.9% |
| Otsu-RGB [16] | 95.2% | 74.6% | 75.0% |
| Otsu-PCA [17] | 77.8% | 99.0% | 98.7% |
| TDLS [18] | 91.6% | 99.1% | 98.7% |
| Proposed Method | **95.9%** | **99.2%** | **99.0%** |

TABLE IV
QUANTITATIVE COMPARISON OF LESION SEGMENTATION RESULTS FOR ALL OF THE IMAGES IN THE DATASET. BEST RESULTS ARE BOLDED.

| Segmentation Algorithm | Segmentation Performance | | |
| --- | --- | --- | --- |
| | Sensitivity | Specificity | Accuracy |
| L-SRM [13] | 89.4% | 92.7% | 99.3% |
| Otsu-R [15] | 87.3% | 85.4% | 84.9% |
| Otsu-RGB [16] | 93.6% | 80.3% | 80.2% |
| Otsu-PCA [17] | 79.6% | **99.6%** | 98.1% |
| TDLS [18] | 91.2% | 99.0% | 98.3% |
| Proposed Method | **95.2%** | 99.0% | **98.7%** |

## V. CONCLUSION

Computerized analysis of skin lesions by images captured by standard cameras is of great interest due to broad availability of such cameras. Meanwhile accurate detection of lesion region in these images is an essential step for precise diagnosis of melanoma. Segmentation process could be a challenging task in the presence of factors such as illumination variation and low contrast. In our method, the images are preprocessed for the segmentation procedure by applying an edge preserving smoothing filter. The enhanced image is fed to a deep learning network using local and global patches. The simultaneous use of local and global patches caused robustness of our method against illumination variation. Also, effective selection of training patches resulted in an accurate detection of lesion's borders. Experimental results showed that the proposed method can achieve a noticeably high accuracy of 98.7% and sensitivity of 95.2% that outperforms other state-of-the-art methods.